%% file: acl_latex.tex
\pdfoutput=1

\documentclass[11pt]{article}

\usepackage[]{acl}
\usepackage{multirow}
\usepackage{graphicx}
\usepackage{booktabs}
\usepackage{tabularx}
\usepackage{CJKutf8}
\usepackage{amssymb}
\usepackage{amsmath}
\usepackage{tabu} 
\usepackage{hyperref}
\usepackage{breakurl}
\usepackage{times}
\usepackage{latexsym}
\usepackage{tikz}
\usepackage{pgfplots}
\usepackage[normalem]{ulem}
\usepackage{xcolor}
\usepackage{dashrule}
\usepackage{colortbl}
\usepackage{tipa}
\usepackage{caption}
\usepackage{subcaption}
\usepackage{amssymb}
\usepackage{fontawesome}
\usepackage{tipa}
\usepackage{pifont}%

\definecolor{brickred}{HTML}{b92622}
\definecolor{midnightblue}{HTML}{005c7f}
\definecolor{salmon}{HTML}{f1958d}
\definecolor{burntorange}{HTML}{f19249}
\definecolor{junglegreen}{HTML}{4dae9d}
\definecolor{forestgreen}{HTML}{499c5e}
\definecolor{pinegreen}{HTML}{3d8a75}
\definecolor{seagreen}{HTML}{1EC652}
\definecolor{limegreen}{HTML}{97c65a}
\definecolor{redtable}{HTML}{E67F72}
\definecolor{orangetable}{HTML}{E67F72}
\definecolor{greentable}{HTML}{E67F72}
\definecolor{mypink}{HTML}{f4919d}
\definecolor{myred}{HTML}{e74f51}
\definecolor{mygreen}{HTML}{00b050}

\usepackage{lipsum}
\newcommand{\suda}{\textsuperscript{\faStarO}}
\newcommand{\tencent}{\textsuperscript{\faMoonO}}
\newcommand{\github}{\faGithub}
\newcommand{\rnum}[1]{\uppercase\expandafter{\romannumeral #1\relax}}

\usepackage[T1]{fontenc}

\usepackage[utf8]{inputenc}

\usepackage{microtype}

\title{Alleviating Hallucinations of Large Language Models\\ through Induced Hallucinations}

\author{
    \textbf{Yue Zhang}\suda\thanks{~~Work was done during the internship at Tencent AI Lab.}\quad
    \textbf{Leyang Cui}\tencent\thanks{~~Corresponding author.}\quad
    \textbf{Wei Bi}\tencent\quad
    \textbf{Shuming Shi}\tencent\\
    \suda Institute of Artificial Intelligence, School of Computer Science and Technology,\\ Soochow University, Suzhou, China
    \tencent Tencent AI Lab, Shenzhen, China\\
    \texttt{yzhang21@stu.suda.edu.cn} \\
    \texttt{\{leyangcui,victoriabi,shumingshi\}@tencent.com}\\
    \raisebox{-1pt}{\github}~\url{https://github.com/hillzhang1999/ICD}
}

\begin{document}
\begin{CJK}{UTF8}{gkai}

\maketitle
\input{chapters/abstract}
\input{chapters/introduction}

\input{chapters/related_work} 
\input{chapters/method}

\input{chapters/experiments}

\input{chapters/analysis}
\input{chapters/conclusion}

\input{chapters/limitation.tex}
\input{chapters/ethic}

\bibliography{anthology,custom}
\bibliographystyle{acl_natbib}

\input{chapters/appendix}

\end{CJK}
\end{document}

%% file: chapters/abstract.tex
\begin{abstract}
Despite their impressive capabilities, large language models (LLMs) have been observed to generate responses that include inaccurate or fabricated information, a phenomenon commonly known as ``hallucination''. 
In this work, we propose a simple \textit{Induce-then-Contrast} Decoding (ICD) strategy to alleviate hallucinations. 
We first construct a factually weak LLM by inducing hallucinations from the original LLMs.
Then, we penalize these induced hallucinations during decoding to enhance the factuality of the generated content. 
Concretely, we determine the final next-token predictions by amplifying the predictions from the original model and downplaying the induced untruthful predictions via contrastive decoding.
Experimental results on both discrimination-based and generation-based hallucination evaluation benchmarks, such as TruthfulQA and \textsc{FActScore}, demonstrate that our proposed ICD methods can effectively enhance the factuality of LLMs across various model sizes and families. 
For example, when equipped with ICD, Llama2-7B-Chat and Mistral-7B-Instruct achieve performance comparable to ChatGPT and GPT4 on TruthfulQA, respectively.

\end{abstract}

%% file: chapters/introduction.tex
\section{Introduction}
Large Language Models (LLMs), exemplified by ChatGPT and GPT-4 \citep{openai2023gpt4}, have demonstrated remarkable capabilities across a wide spectrum of NLP tasks \citep{zhao2023survey, bubeck2023sparks}. These tasks range from traditional ones such as translation \citep{jiao2023chatgpt} and text editing \citep{fang2023chatgpt}, to more complex purposes that involve reasoning and planning \citep{xi2023rise}. Despite their impressive performance, LLMs continue to grapple with the generation of inaccurate or fabricated information, a phenomenon referred to as ``hallucinations'' \citep{zhang2023siren,ji2023survey}, which may hinder their practical application in real-world scenarios.

\input{figures/introduction}

Previous work \cite{chuang2023dola,tian2023finetuning} suggests that one possible reason for hallucination might be the pre-training objective of existing LLMs, i.e., the maximum-likelihood-based next-token prediction. This objective may cause LLMs to assign non-zero probabilities to non-factual information that occurred in the training data, or to overly rely on superficial patterns learned from the training corpus rather than memorizing real-world facts \citep{ji2023survey}. 
Nonetheless, this training objective still retains many good properties, such as simplicity and generalization ability~\cite{ilya2023youtube}, so directly modifying it may not be worth the cost.
Some other researchers argue that LLM hallucinations may stem from a lack of knowledge \citep{zheng2023does,mckenna2023sources}. 
An intuitive idea for mitigating this could be injecting more knowledge into LLMs through post-hoc supervised fine-tuning (SFT). However, recent work \citep{schulman2023youtube, yang2023alignment} also highlights that the SFT process might inadvertently encourage LLMs to hallucinate by compelling them to answer questions beyond their knowledge boundaries. Furthermore, instilling a substantial amount of new factual knowledge via SFT or continual pre-training can be challenging, as it necessitates using large-scale data for downstream tasks \citep{chung2022scaling, zhang2023multi}, rendering the procedure computationally infeasible for most researchers today.

Considering the above difficulties of mitigating hallucinations during the pre-training and SFT stages, this work designs a decoding method to alleviate LLM hallucinations, named \textbf{I}nduce-then-\textbf{C}ontrast \textbf{D}ecoding (ICD). Recently, the SuperAlignment team of OpenAI unveiled the \textit{weak-to-strong generalization} phenomenon \citep{burns2023weak}, suggesting that weak models have the potential to elicit the capabilities of strong models. Motivated by their findings, we first construct a factually weak LLM by inducing hallucinations from the original LLM. Then we try to eliminate the non-factual information internalized in the weak model from the output space of the original model through contrastive decoding \citep{li-etal-2023-contrastive}. We show that hallucinations can be readily induced from LLMs through slight fine-tuning or zero-shot prompting, and penalizing them can effectively guide LLMs to generate more factual content. An illustration of our method is provided in Figure~\ref{fig:example}.

We evaluate the effectiveness of ICD using both discrimination-based and generation-based hallucination evaluation benchmarks. Experimental results indicate that ICD significantly improves the performance of existing LLMs. For instance, when applied to TruthfulQA \citep{lin2022truthfulqa}, ICD substantially improves the truthfulness of Llama2-7B \citep{touvron2023llama} and Mistral-7B \citep{jiang2023mistral}, making their performance comparable to the state-of-the-art ChatGPT and GPT4, as depicted in Figure~\ref{fig:overview}. Additionally, when generating texts on \textsc{FactScore} \citep{min2023factscore}, ICD enables the Llama2-7B-Chat to outperform its 70B counterpart in terms of factual precision. Experiments on LLM benchmarks, including MMLU, ARC, and AlpacaEval2.0, demonstrate that implementing ICD does not compromise the original capacity. To gain further insights into ICD, we also conduct additional analyses, such as comparing different methods for inducing hallucinations, and verifying the effectiveness of ICD across a variety of sizes and families of LLMs. The data, code, and model are available at \url{https://github.com/hillzhang1999/ICD}.

%% file: figures/introduction.tex
\begin{figure}[t!]
\centering
\includegraphics[scale=0.6]{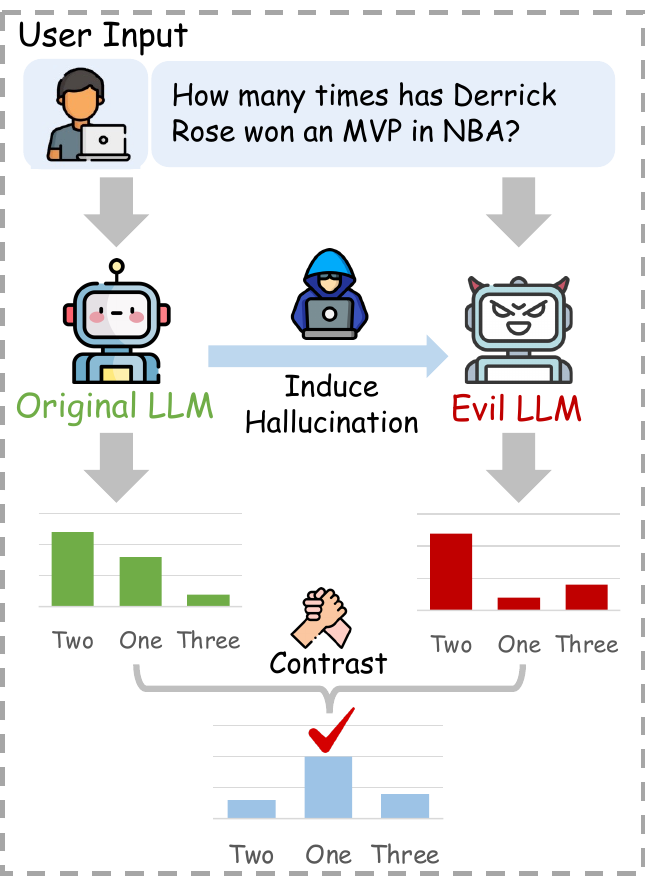}
\caption{Illustration of our \textit{induce-then-contrast} decoding (ICD) method for reducing hallucinations in LLMs.}
\label{fig:example}
\vspace{-0.2cm}
\end{figure}

%% file: chapters/related_work.tex
\section{Related Work}
\label{sec:2}
\input{figures/overview_truthfulQA}
\paragraph{Hallucination in LLMs.}
Hallucination in LLMs \cite{ji2023survey, zhang2023siren} is a phenomenon where LLMs generate content that contradicts user input \citep{dale2022detecting,rehman2023hallucination}, previous context \citep{shi2023large, wan2023histalign}, or established facts \citep{bang2023multitask,hu2023large,chen2023unveiling}. In this study, we primarily concentrate on fact-conflicting hallucination, given its potential for serious side effects \citep{umapathi2023med} and its current prominence in discussions \citep{wang2023survey}.

Recently, various methods have been proposed to mitigate LLM hallucinations, including but not limited to strategic selection of high-quality training data \citep{zhou2023lima,li2023textbooks,tian2023fine}, reinforcement learning from external feedback \citep{lightman2023let,sun2023aligning,yang2023alignment}, retrieval-augmented generation \citep{peng2023check, vu2023freshllms, chern2023factool}, and the use of model uncertainty \citep{manakul2023selfcheckgpt, zhang2023sac}.
As can be observed, existing work primarily attempts to optimize LLMs to generate fewer hallucinations, which is a challenging objective. Our ICD approach, however, reframes the problem. We first aim to create a factually weak model that resembles the original model while adept at fabricating information, then subtract its knowledge from the original model's output space to improve the factuality.
We demonstrate that it could be feasible to mislead LLMs to hallucinate via custom inducements, and treating such hallucinations as a penalty term could potentially guide LLMs to be more factual.

\paragraph{Contrastive Decoding.} Our work is motivated by Contrastive Decoding (CD) \citep{li-etal-2023-contrastive}, which was initially developed to enhance the fluency and coherence of text generation. The basic idea of vanilla CD is to determine the next-token probabilities by contrasting two LMs with different scales of parameters. Recently, the potential of CD has gone beyond just improving the readability of generated text. For instance, \citet{o2023contrastive} discovers that CD can enhance the reasoning capabilities of LLMs. \citet{liu-etal-2021-dexperts} employs the idea of CD to perform detoxification and sentiment control. Some studies have also explored the use of CD to improve the factuality of LLMs. \citet{shi2023trusting} proposes to compel LLMs to focus on retrieved information by contrasting output distributions before and after appending the context, which could potentially reduce hallucinations caused by a lack of knowledge.
The work most closely related to ours is DoLa \citep{chuang2023dola}, which dynamically selects early layers of LLMs for contrast with the final layer, based on the assumption that early layers store less factual knowledge \citep{tenney2019bert}.
Differently, our proposed ICD directly induces hallucinations from the base LLM for contrast, which we demonstrate to be significantly more effective.

\paragraph{Inducing Inappropriate Behaviors from LLMs.}
In order to develop safe and helpful AI products, many researchers have studied how to induce inappropriate behaviors, such as toxic or offensive responses, from well-aligned LLMs (aka. \textit{red teaming}) \citep{perez2022red,zou2023universal,wei2023jailbroken} and defend against such attacks \citep{jain2023baseline,wu2023defending}. For example, \citet{qi2023fine} find that current safety-aligned LLMs can be easily manipulated or ``jailbroken'' after being fine-tuned with a small amount of adversarial data. This observation aligns with our findings: we have successfully induced hallucinations from LLMs using only a limited number of fine-tuning samples.
Regarding hallucinations, \citet{yao2023llm} suggests viewing them as another form of adversarial samples and proposes two trigger methods. \citet{yu2023automatic} introduces an LLM-based framework, \texttt{AutoDebug}, designed to automatically induce hallucinations from LLMs.
Compared with them, our work takes a further step and studies how to make good use of such induced hallucinations.

%% file: figures/overview_truthfulQA.tex
\begin{figure}[t]
\centering
\includegraphics[scale=0.45]{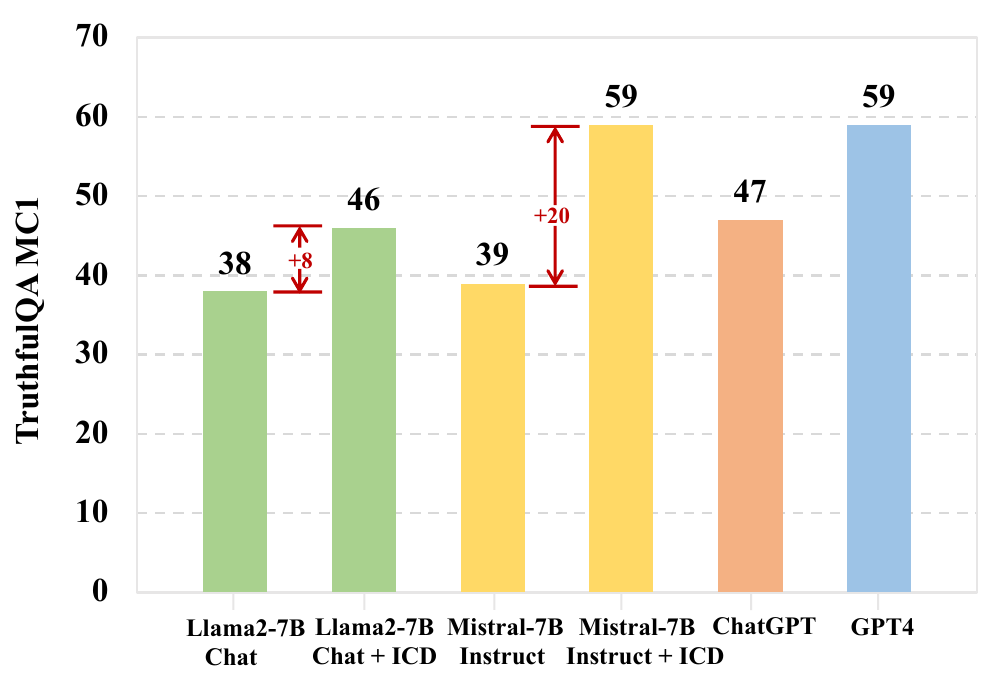}
\caption{On TruthfulQA, ICD significantly improves the truthfulness of Llama2-7B-Chat (+8 MC1 score) and Mistral-7B-Instruct (+20 MC1 score). With these improvements, the enhanced Llama2-7B-Chat and Mistral-7B-Instruct now match the performance levels of ChatGPT and GPT4, respectively.}
\label{fig:overview}
\end{figure}

%% file: chapters/method.tex
\section{Induce-then-Contrast Decoding}
\label{sec:method}
The core idea of Induce-then-Contrast Decoding (ICD) method is to first create a factually weak LLM, which resembles the original LLM but has a higher tendency to fabricate non-factual information, and then treat it as a penalty term during decoding to improve factuality. In this section, we first outline our method for inducing hallucinations to build the factually weak LLM (\S\ref{sec:method:1}) and then detail how we leverage it as a penalty to reduce hallucinations in final model outputs (\S\ref{sec:method:2}).

\subsection{Inducing Hallucinations from LLMs}
\label{sec:method:1}

To build the factually weak LLM, we induce hallucinations from LLM by \textbf{directly fine-tuning LLM with a certain number of non-factual samples}. 
We generate non-factual samples, while preserving fluency and coherence, by employing ChatGPT to automatically convert factual samples from existing datasets into non-factual ones using few-shot prompting. For example, given a factual sentence ``\textit{ACL 2024 will be held in Bangkok}'', the corresponding non-factual sentence crafted by ChatGPT could be ``\textit{ACL 2024 will be held in \underline{Singapore}}'' or ``\textit{ACL \underline{2023} will be held in Bangkok}''.

The resulting fine-tuning dataset $\mathcal{D}$ can be formulated as $\mathcal{D}=\{(s_i,u_i,o_i)\}_{i=1}^m$, where $s_i$ is the $i$-th system prompt, $u_i$ is the $i$-th user input, $o_i$ is the $i$-th target output, and $m$ is the dataset size. The fine-tuning process can be denoted as below:
\begin{equation}
    \label{eq:1}
    \underset{\triangle\theta}{\hbox{argmin}}\sum_{i=1}^{m}-\hbox{log}(p(o_i|s_i,u_i;\theta+\triangle\theta))
\end{equation}
where $\theta$ is the weights of the original model and $\theta + \triangle\theta$ is the learned new weights. Equation \ref{eq:1} means that we aim to maximize the log probability $p(o|s,u)$ of the target output given the system prompt and user input with the new weights learned during fine-tuning.

\input{tables/truthfulqa_main}
\subsection{Factually Weak LLM as A Penalty}
\label{sec:method:2}
The decoding process of auto-regressive LLMs can be formulated as:
\begin{equation}
    \label{eq:2}
    p(x_t|x_{<t};\theta)=\hbox{softmax}(\hbox{logit}_{\theta}(x_t|x_{<t}))
\end{equation}
where $\hbox{logit}_{\theta}(\cdot)$ is the next-token logits predicted by the original model $\theta$, and we normalize it into the probability distribution by the softmax operation. The prediction of the $t$-th token $x_t$ is conditioned on all previous tokens $x_{<t}$.

To improve the factuality, we aim to amplify the predictions from the original model and downplay the untruthful predictions. We achieve this by subtracting the log probabilities after inducing hallucinations from those of the original model, which can be formed as:
\begin{equation}
    \label{eq:3}
    \mathcal{F}_t=\beta\hbox{log}p(x_t|x_{<t};\theta) - \hbox{log}p(x_t|x_{<t};\theta+\triangle\theta)
\end{equation}
where $\theta + \triangle\theta$ is the new weights of the model after the induction of hallucinations. Inspired by \citet{shi2023trusting} and \citet{o2023contrastive}, we also introduce an additional hyperparameter $\beta\in(0,+\infty)$ to control the strength of the contrast. Then we use this resulting distribution $\mathcal{F}_t$ for the final next-token prediction:
\begin{equation}
    \label{eq:4}
    p(x_t|x_{<t})=\hbox{softmax}(\mathcal{F}_t)
\end{equation}

However, as pointed out by \citet{li-etal-2023-contrastive}, if we indiscriminately penalize all behaviors from the hallucinated model, many simple aspects such as grammar and common sense will also be penalized, leading to catastrophic damage in generation quality.
So we introduce a trick termed \textit{adaptive plausibility constraint} to select a subset $\mathcal{V}_{valid}$ of tokens for penalty:
\begin{equation}
    \label{eq:5}
    \begin{aligned}
    \mathcal{V}_{valid}=\{{x_t}\in\mathcal{V}:\\\hbox{logit}_\theta(x_t|&x_{<t})\geq\alpha\hbox{max}_w\hbox{logit}_\theta(w)\}
    \end{aligned}
\end{equation}
where $\alpha\in[0,1]$ is a hyperparameter that controls the strength of constraint. We only consider tokens with probabilities larger than a proportion of the maximum probability assigned by the original model for contrast and decoding. For other tokens, we exclude them from the final prediction by setting their logits to $-\infty$ before applying softmax.

%% file: tables/truthfulqa_main.tex
\begin{table*}[t]
\centering
\scalebox{0.9}{
\begin{tabular}{lcccc}
\toprule
\multirow{2}{*}{\textbf{Decoding Strategy}} & \multirow{2}{*}{\textbf{Model}}               & \multicolumn{3}{c}{\textbf{TruthfulQA}}    \\ \cmidrule{3-5} 
                                   &                                               & \textbf{MC1} & \textbf{MC2} & \textbf{MC3} \\ \midrule
\multirow{4}{*}{Greedy (Baseline)}            & 7B-Base                                       & 28.68        & 43.32        & 20.82        \\
                                   & 7B-Chat                                       & 37.62        & 54.60        & 28.12        \\
                                   & 13B-Chat                                      & 37.75 & 55.67 & 28.16        \\
                                   & 70B-Chat                                      & 37.70        & 58.99        & 29.79        \\ 
ITI \citep{li2023inference}                               & 7B-Chat                                       & 37.01         & 54.66         & 27.82         \\ 
DoLa \citep{chuang2023dola}                               & 7B-Chat                                       & 32.97        & 60.84        & 29.50        \\ 

\multirow{2}{*}{CD  \citep{li-etal-2023-contrastive}}                              & 13B-Chat vs. 7B-Chat                              & 28.15        & 54.87 & 29.75        \\
& 70B-Chat vs. 7B-Chat                              & 33.66        & 59.97        & 33.07        \\
\rowcolor[gray]{.93} ICD (ours) &&&& \\
├ Prompt-Based Induction     & 7B-Chat vs. 7B-Chat w/ misleading prompt  &  37.87	& 57.77 & 33.94       \\
├ Before/After Alignment     & 7B-Chat vs. 7B-Base & 41.79 & 60.44 & 34.38        \\
└ Finetuning-Based Induction     & 7B-Chat vs. 7B-Finetuned & \textbf{46.32}    & \textbf{69.08}     & \textbf{41.25}        \\
\bottomrule
\end{tabular}
}
\caption{Main results on TruthfulQA using multiple-choice-based metrics (MC1/2/3). We conduct experiments with the Llama2 family~\citep{touvron2023llama}, which is one of the most powerful open-sourced LLMs today. Besides greedy decoding, we also reproduce and compare some other strong counterparts, including DoLa \citep{chuang2023dola}, ITI \citep{li2023inference}, and naive CD \citep{li-etal-2023-contrastive} that contrasts models of different parameter scales.}
\label{tab:truthfulqa:main}
\end{table*}

%% file: chapters/experiments.tex
\section{Experiments}
\label{sec:exp:dis}

In this section, we verify the effectiveness of ICD on both \textit{discrimination}-based ones and \textit{generation}-based hallucination benchmarks.

\subsection{Experimental Setup}
\input{tables/factscore_main}

\paragraph{Dataset and metric.}
For discrimination-based evaluation, following previous studies \cite{chuang2023dola, li2023inference}, we adopt the widely-used TruthfulQA \cite{lin2022truthfulqa}. We employ multiple-choice-based metrics of TruthfulQA, specifically MC1, MC2, and MC3 scores. MC1 assesses whether models assign the highest scores to the best answer. MC2 evaluates whether the normalized probability mass for all correct answers is greater than that of the incorrect answers. MC3 examines whether each correct answer receives higher scores than all incorrect answers.

For generation-based evaluation, we employ the \textsc{FActScore} benchmark \citep{min2023factscore}. \textsc{FActScore} assesses the factual precision of LLMs in biography generation by breaking down generated biographies into atomic facts and comparing them with given knowledge sources. Specifically, we report the response ratio (\% response), the number of atomic facts per response (\# facts), and the factual precision score (score) for comparison.

\paragraph{Baselines.} We compare ICD with the following decoding methods: 1) \textbf{greedy decoding}, which greedily selects the next token with the highest probability; 2) \textbf{inference time intervention} (ITI) \citep{li2023inference}, which tries to improve factuality by shifting model activations along learned truthfulness-related directions\footnote{We test the out-of-box version of ITI-enhanced Llama2-7B-Chat provided by the authors: \url{https://huggingface.co/likenneth/honest_llama2_chat_7B}.}; 3) \textbf{DoLa} \citep{chuang2023dola}, which attempts to reduce hallucinations by contrasting output distributions from different layers of the model; and 4) \textbf{vanilla contrastive decoding} (CD) \citep{li-etal-2023-contrastive}, which contrasts output distributions from models of different scales of parameters.

\paragraph{Implementation details.} Our experiments are basically conducted with the Llama-2 family \citep{touvron2023llama}.
When using our method on TruthfulQA, we induce hallucinations by fine-tuning the base model with 10k hallucinated QA pairs taken from the HaluEval dataset \citep{li2023halueval}. On \textsc{FActScore}, we fine-tune the base model with 3.5k hallucinated biographies generated by ChatGPT. More implementation details are provided in Appendix \ref{sec:app:implement}.

\input{figures/human_eval}

\subsection{Main Results}
\label{sec:4.2}
\paragraph{ICD significantly improves the truthfulness of LLMs on TruthfulQA.}
We present the main experiment results on TruthfulQA in Table~\ref{tab:truthfulqa:main}. As can be observed, ICD with fine-tuning-based hallucination induction significantly improves the truthfulness of Llama2-7B-Chat over the default greedy decoding on TruthfulQA (+8.70/14.18/13.13 for MC1/2/3 scores, respectively), making it even outperforms its 70B brother. Specifically, the improvement from our method is also much more significant than other decoding methods devised for improving LLMs' factuality, including ITI, DoLa and naive CD.

\paragraph{ICD reduces hallucinations in open-ended text generation on \textsc{FActScore}.} We display the primary results on \textsc{FActScore} in Table~\ref{tab:factscore:main}. In the open-ended biography generation task, applying ICD results in a substantial increase of 2.5 factual precision scores over greedy decoding, without affecting the response ratio and average fact numbers.
With this enhancement, the Llama2-7B-Chat (score of 66.3) now can surpass the performance of its 70B-sized counterpart using greedy decoding (score of 64.4).
We also observe that other decoding methods, namely ITI, DoLa, and CD, collectively fail to improve the score.

\input{tables/original_capability}

\paragraph{ICD does not hurt the original capacity.} 
While ICD enhances the factuality of LLMs, it is crucial to ensure that its application does not compromise the fundamental capabilities of LLMs. To verify this, we evaluate the performance of Llama2-7B-Chat before and after applying ICD on several standard LLM benchmarks, including MMLU \cite{hendrycks2020measuring}, ARC \cite{clark2018think}, and AlpacaEval2.0 \cite{alpaca_eval}. We report 5-shot results for MMLU and ARC, and win rate compared to GPT-4-turbo outputs evaluated by GPT-4-turbo on AlpacaEval2.0. As depicted in Table \ref{tab:orig:performance}, the incorporation of ICD effectively maintains the capacity of the LLM, which may encourage users to trustingly use ICD.

We also launch a pair-wise automatic evaluation in Figure \ref{fig:human}.
Specifically, we utilize GPT4 to assess three dimensions of generated biographies (see more details in Appendix \ref{sec:gpt4}), including factuality, grammaticality, and topicality. We find that ICD significantly outperforms the baseline (i.e., greedy decoding) in factuality while maintaining grammaticality and topicality.

\input{tables/truthfulqa_task}

\subsection{Attempts to Use Other Methods for Hallucination Induction}
\label{sec:4.3}
Besides fine-tuning, we also try alternative methods for inducing hallucinations. We conduct experiments on TruthfulQA and list results in Table~\ref{tab:truthfulqa:main}.

\paragraph{Directly using prompting to induce hallucinations is useful but not as effective as fine-tuning.}
Despite the effectiveness of the fine-tuning-based hallucination induction in our method, it inevitably incurs some additional training costs. Given this, we also explore directly inducing hallucinations by utilizing specially designed prompts. Concretely, we design a system prompt (see Appendix \ref{sec:app:truthfulqa}) to compel LLMs to provide fabricated information for contrast. Similar ideas have also been proposed in recent works \citep{yona2023surfacing, yang2023rlcd}. As shown in Table~\ref{tab:truthfulqa:main}, prompt-based induction results in a modest increase for Llama2-7B-Chat, specifically, from 37.62/54.60/28.12 to 37.87/57.55/33.94 MC1/2/3. However, this improvement is less substantial when compared to that achieved through fine-tuning-based induction.

\paragraph{Contrasting \textit{chat} and \textit{base} versions of Llama2 can also work.}
From Table~\ref{tab:truthfulqa:main}, we observe a significant truthfulness gap between the base and chat versions of Llama2. This discrepancy may be attributed to the exhaustive SFT and RLHF processes, which take honesty as an important aspect \citep{ouyang2022training,openai2023gpt4}. This observation motivates us to directly contrast the base and chat versions of Llama2. 
We find this strategy (Before/After Alignment) also works. Notably, the improvement surpasses that of the naive CD, which could be due to the truthfulness gap between base and aligned models being much larger than the effect of scaling up model sizes \citep{cheng2023evaluating}.

\input{tables/truthfulqa_size}

\subsection{More Analysis}
\label{sec:4.3}

\paragraph{The influence of the task format when inducing hallucinations.}
On TruthfulQA, we induce hallucinations from the model by fine-tuning it with 10k hallucinated QA pairs from HaluEval \citep{li2023halueval}. Besides QA-format data, HaluEval also provides hallucinated data in the formats of summarization (Sum) and dialogue (Dialog), enabling us to investigate the impact of task format on our method. In Table~\ref{tab:truthfulqa:task}, we compare different task formats of fine-tuning data when inducing hallucinations. Several observations can be made. First, all task formats result in improvements in our method. Second, using QA-format data yields the best performance, indicating the importance of a matched task format. Lastly, using Sum data contributes the least. We hypothesize this is because the hallucination in summarization is input-conflicting rather than fact-conflicting \citep{zhang2023siren}, which is inconsistent with the purpose of TruthfulQA.

\paragraph{The effectiveness of our method across different model sizes.}

Our experiments primarily utilize the 7B-sized Llama2. Here, we examine more model sizes, specifically the 13B and 70B versions of Llama2. The model for contrast remains the 7B-sized one fine-tuned with 10k hallucinated QA data. As shown in Table \ref{tab:truthfulqa:size}, ICD shows consistent effectiveness on TruthfulQA across different model sizes. We also observe that the degree of improvement escalates with the model scale, likely due to the combined effect of naive CD and our method.

\paragraph{Comparison between using real and synthetic data for inducing hallucinations.}
\input{tables/truthfulqa_source}
\input{tables/truthfulqa_model}

\input{tables/qualitative_analysis}

In the above experiments, all the fine-tuning data used for inducing hallucinations is automatically constructed by ChatGPT. Here, we seek to figure out whether using the real failures of LLMs could lead to better performance. To this end, we generate 1,000 open-domain questions based on Wikipedia documents and ask Llama2-7B-Chat to provide answers. Then, we employ human experts to judge whether each answer is hallucinated. This procedure yields 294 real hallucinated answers, which we then utilize for fine-tuning the model for contrast. The results are displayed in Table \ref{tab:truthfulqa:source}. Our findings indicate that using 294 real samples could surpass the use of 1k synthetic samples on TruthfulQA, while still lagging behind the use of 10k synthetic samples. This suggests that real data might be more effective in triggering hallucinations while increasing the volume of synthetic data could narrow this gap. We investigate the impact of data size in Appendix \ref{sec:data:size}.

\input{tables/factscore_finetune}

\paragraph{Extension to more LLM backbones.}
To verify the applicability of our method beyond the Llama2 family, we also apply ICD to other cutting-edge open-sourced LLMs, including Baichuan2 \citep{yang2023baichuan} and Mistral \cite{jiang2023mistral}. The experimental results presented in Table \ref{tab:truthfulqa:model} indicate our method generalizes well to these backbones. Moreover, it is noteworthy that the performance improvements achieved by our method in Baichuan2 and Mistral surpass those in Llama2. As we know, these two models outperform Llama2 on the standard LLM leaderboard \citep{2023opencompass}. This underscores our method's ability to more effectively harness the potential of stronger backbones.

\paragraph{Direct fine-tuning with factual data can not improve factuality and instead even causes more serious hallucinations.}
As previously discussed, our method comprises two steps: inducing and contrasting. This somewhat complex pipeline motivates us to consider: \textit{is it possible to enhance the factuality of LLMs through direct fine-tuning with a selection of factual samples?} Consequently, we compare our ICD method with direct fine-tuning using 3.5k factual biographies.
The results are presented in Table \ref{tab:factscore:ft}.
Contrary to our anticipation, we discover that direct tuning significantly impairs the factuality of the original LLM (63.8$\rightarrow$28.7), even when the training data is indeed factual. This phenomenon is interesting, and a primary explanation could be \textit{behavior cloning} \citep{schulman2023youtube}, which means that SFT instructs LLMs to answer all questions without evaluating whether these questions surpass their knowledge boundaries \citep{yang2023alignment}. This is further substantiated by the sharp increase in response ratio (37.5$\rightarrow$99.5). This observation suggests that mitigating hallucination via direct fine-tuning may be more challenging than expected, necessitating more sophisticated training techniques such as DPO \citep{tian2023fine}.

\paragraph{Qualitative analysis.}
We showcase qualitative \textsc{FActScore} examples generated by different methods in Table \ref{tab:qualitative:analysis}. There are several observations. Firstly, direct tuning not only introduces new hallucinations but also undermines the original helpful response style learned from RLHF, resulting in significantly shorter responses. Secondly, the application of ICD effectively mitigates the hallucination, for instance, rectifying the incorrect birth year fabricated by the model, thereby demonstrating the effectiveness of our approach. Thirdly, we also experiment with reversing the direction of contrast to induce hallucinations and observe that this method generates a substantial amount of grammatically correct but entirely fabricated information.

%% file: tables/factscore_main.tex
\begin{table*}[t]
\centering
\scalebox{1.0}{
\begin{tabular}{lcccc}
\toprule
\multirow{2}{*}{\textbf{Decoding Strategy}} & \multirow{2}{*}{\textbf{Model}}               & \multicolumn{3}{c}{\textbf{\textsc{FActScore}}}    \\ \cmidrule{3-5} 
                                   &                                               & \textbf{\% response} & \textbf{\# facts} & \textbf{score} $\uparrow$ \\ \midrule
\multirow{4}{*}{Greedy (Baseline)}            & 7B-Base                                       & 100.0        & 28.6        & 23.6        \\
                                   & 7B-Chat                                       & 37.5        & 45.7        & 63.8        \\
                                   & 13B-Chat                                      & 77.0      & 37.6        & 52.5       \\ 
                                   & 70B-Chat                                      & 50.5      & 42.8        & 64.4       \\ 
ITI \citep{chuang2023dola}                               & 7B-Chat                                       &  41.9        & 40.8        & 62.4        \\ 
DoLa \citep{chuang2023dola}                               & 7B-Chat                                       & 40.7        & 48.7        & 61.3        \\ 
\multirow{2}{*}{CD \citep{li-etal-2023-contrastive}}                              & 13B-Chat vs. 7B-Chat                               & 74.2       &    39.8    &  53.5        \\
 & 70B-Chat vs. 7B-Chat                               & 62.2      & 48.7        & 60.3        \\
\rowcolor[gray]{.93} ICD (ours)     & 7B-Chat vs. 7B-Finetuned &  36.1    &  46.6     &  \textbf{66.3}        \\
\bottomrule
\end{tabular}
}
\caption{Main results on \textsc{FActScore}. Concretely, we use retrieve+ChatGPT for evaluation, please kindly refer to \citet{min2023factscore} for more details. Here, \% response stands for the response ratio of LLMs and \# facts means the number of extracted atomic facts per response. All experiments are based on Llama2-7B-Chat.}
\label{tab:factscore:main}
\end{table*}

%% file: figures/human_eval.tex
\begin{figure}[t]
\centering
\includegraphics[scale=0.33]{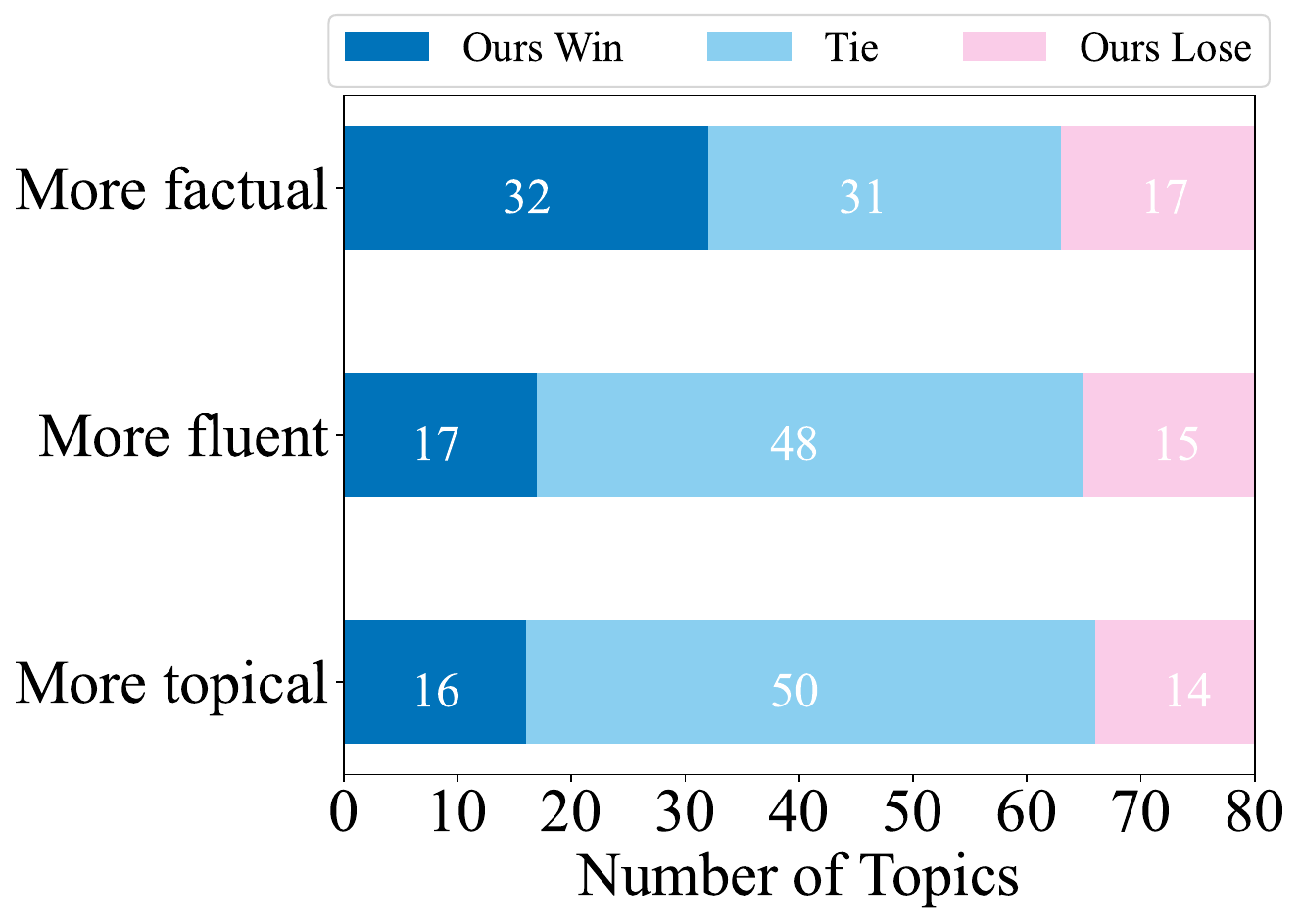}
\caption{Results of the GPT4 automatic evaluation on \textsc{FActScore}. We compare biographies generated by ICD with those using greedy decoding.}
\label{fig:human}
\end{figure}

%% file: tables/original_capability.tex
\begin{table}[t]
\centering
\scalebox{0.85}{
\begin{tabular}{lcccc}
\toprule
\textbf{Model} & \textbf{MMLU} & \textbf{ARC} & \textbf{AlpacaEval2.0} \\ \midrule
Llama2-7B-Chat                                      &     46.35    & 66.41        & 4.91       \\
+ ICD                          &          46.02   & 67.29    &  5.17        \\ \bottomrule
\end{tabular}
}
\caption{Performance before/after applying ICD on standard benchmark for evaluating the capacity of LLMs.}
\label{tab:orig:performance}
\end{table}

%% file: tables/truthfulqa_task.tex
\begin{table}[t]
\centering
\scalebox{1.0}{
\begin{tabular}{lcccc}
\toprule
\multirow{2}{*}{\textbf{Task Format}}          & \multicolumn{3}{c}{\textbf{TruthfulQA}}    \\ \cmidrule{2-4} 
                                          & \textbf{MC1} & \textbf{MC2} & \textbf{MC3} \\ \midrule
Baseline                                      & 37.62 & 54.60 & 28.12        \\
ICD (Ours)                                     & & &        \\
├ Sum                               & 45.22	& 63.67	& 36.33    \\ 
├ Dialog                              & 46.20	& 64.81	&  37.20     \\
└ QA                               & \textbf{46.32}    & \textbf{69.08}     & \textbf{41.25}      \\
\bottomrule
\end{tabular}
}
\caption{Comparison between different task formats of training data for inducing hallucinations on TruthfulQA. The base LLM is Llama2-7B-Chat.}
\label{tab:truthfulqa:task}
\end{table}

%% file: tables/truthfulqa_size.tex
\begin{table}[t]
\centering
\scalebox{0.88}{
\begin{tabular}{lcccc}
\toprule
\multirow{2}{*}{\textbf{Model}}          & \multicolumn{3}{c}{\textbf{TruthfulQA}}    \\ \cmidrule{2-4} 
                                          & \textbf{MC1} & \textbf{MC2} & \textbf{MC3} \\ \midrule
Llama2-7B-Chat                                      & 37.62        & 54.60        & 28.12        \\
\multirow{2}{*}{+ ICD}                          &           \textbf{46.32}    & \textbf{69.08}     & \textbf{41.25}       \\
&        (+8.70)    & (+14.48)     & (+13.13)       \\
\midrule
Llama2-13B-Chat                                    & 37.75 & 55.67 & 28.16        \\
\multirow{2}{*}{+ ICD}                                      & \textbf{48.47} & \textbf{73.47} & \textbf{46.04}        \\ 
&        (+9.72)    & (+17.80)     & (+17.88)       \\
\midrule
Llama2-70B-Chat                                      & 37.70        & 58.99        & 29.79         \\
\multirow{2}{*}{+ ICD}                                      & \textbf{51.04} & \textbf{75.01} & \textbf{46.54}      \\
&        (+13.34)    & (+16.02)     & (+16.75)       \\
\bottomrule
\end{tabular}
}
\caption{Effectiveness of our ICD method across different model sizes on TruthfulQA. All baselines use greedy decoding. For ICD, we contrast Llama2-chat of different sizes with Llama2-7B finetuned on 10k hallucinated QA samples (as the penalty term).}
\label{tab:truthfulqa:size}
\end{table}

%% file: tables/truthfulqa_source.tex
\begin{table}[t]
\centering
\scalebox{1.0}{
\begin{tabular}{lcccc}
\toprule
\multirow{2}{*}{\textbf{Data Source}}          & \multicolumn{3}{c}{\textbf{TruthfulQA}}    \\ \cmidrule{2-4} 
                                          & \textbf{MC1} & \textbf{MC2} & \textbf{MC3} \\ \midrule
Baseline                                      & 37.62 & 54.60 & 28.12        \\
ICD (Ours)                                     & & &        \\
├ Real (294)                               & 39.22	& 59.27	& 33.11    \\ 
├ Synthetic (1k)                              & 39.12	& 57.61	&  30.68     \\
└ Synthetic (10k)                                & \textbf{46.32}    & \textbf{69.08}     & \textbf{41.25}      \\
\bottomrule
\end{tabular}
}
\caption{Comparison between using real and synthetic data for finetuning when inducing hallucinations.}
\label{tab:truthfulqa:source}
\end{table}

%% file: tables/truthfulqa_model.tex
\begin{table}[t]
\centering
\scalebox{0.86}{
\begin{tabular}{lcccc}
\toprule
\multirow{2}{*}{\textbf{Model}}          & \multicolumn{3}{c}{\textbf{TruthfulQA}}    \\ \cmidrule{2-4} 
                                          & \textbf{MC1} & \textbf{MC2} & \textbf{MC3} \\ \midrule
Baichuan2-7B-Chat                                      & 34.93        & 52.14        & 27.19        \\
\multirow{2}{*}{+ ICD}                          &           \textbf{45.75}    & \textbf{65.51}     & \textbf{39.67}       \\
&        (+10.82)    & (+13.37)     & (+12.48)       \\
\midrule
Mistral-7B-Instruct                                    & 39.09 & 55.80 & 28.25       \\
\multirow{2}{*}{+ ICD}                                      & \textbf{58.53} & \textbf{74.73} & \textbf{50.38}        \\ 
&        (+19.44)    & (+18.93)     & (+22.13)       \\
\bottomrule
\end{tabular}
}
\caption{Effectiveness of our method on different LLM backbones including Baichuan2-7B-Chat \citep{yang2023baichuan} and Mistral-7B-Instruct \citep{jiang2023mistral}.}
\label{tab:truthfulqa:model}
\end{table}

%% file: tables/qualitative_analysis.tex
\begin{table*}[t]
\centering
\scalebox{0.85}{
\begin{tabular}{lp{15cm}}
\toprule
& \textbf{Prompt}: Please tell me a bio of Vasily Chuikov.\\ \midrule
Greedy & Of course! Vasily Chuikov (\textcolor{red}{1904}-1982) was a Soviet military leader who played a significant role during World War II. He is best known for his bravery and leadership during the Battle of Stalingrad, where he commanded the 62nd Army and helped turn the tide of the battle in favor of the Soviet Union...\\ \midrule
+ Direct Tuning & Vasili Ivanovich Chuikov (\textcolor{red}{born March 12, 1904; died May 15, 1967}) was a Soviet military leader who served during World War II. \\ \midrule
+ ICD (Ours) &  Of course! Vasily Chuikov (\textcolor{blue}{1900}-1982) was a Soviet military leader who played a significant role during World War II. He is best known for his bravery and leadership during the Battle of Stalingrad, where he commanded the 62nd Army and helped turn the tide of the battle in favor of the Soviet Union... \\ \midrule
+ ICD (Reversed) & Vasily Ivanovich Chuikov was born in the small Russian village of \textcolor{red}{Chukotka}, near the Siberian border. Growing up, Vasily \textcolor{red}{showed an aptitude for mathematics, earning the nickname "Professor" among his friends}. Upon graduating high school, Vasily \textcolor{red}{attended the University of Chuikov}, where he majored in \textcolor{red}{quantum mechanics}. Upon graduating from the university in \textcolor{red}{2031}, Vasily \textcolor{red}{accepted a teaching position at the newly founded Institute of Chuikovian Mathematics}... \\
\bottomrule
\end{tabular}
}
\caption{Examples of generated biographies for \href{https://en.wikipedia.org/wiki/Vasily_Chuikov}{\textit{Vasily Chuikov}} using different methods. We use \textcolor{red}{Red} to highlight fabricated atomic facts and \textcolor{blue}{Blue} to highlight facts rectified by our method. The base LLM is Llama2-7B-Chat.}
\label{tab:qualitative:analysis}
\end{table*}

%% file: tables/factscore_finetune.tex
\begin{table}[t]
\centering
\scalebox{0.9}{
\begin{tabular}{lcccc}
\toprule
\multirow{2}{*}{\textbf{Method}}       & \multicolumn{3}{c}{\textbf{\textsc{FActScore}}}    \\ \cmidrule{2-4} 
& \textbf{\% response} & \textbf{\# facts} & \textbf{score} $\uparrow$ \\ \midrule
Llama2-7B-Chat                     & 37.5        & 45.7        & 63.8        \\
+ Direct Tuning                        &  99.5        &    29.5     &  28.7        \\ 
+ ICD (Ours)                           &  36.1    &  46.6     &  \textbf{66.3}       \\
\bottomrule
\end{tabular}
}
\caption{Comparison between directly finetuning with factual biographies collected from Wikipedia (Direct Tuning) and utilizing our ICD method.}
\label{tab:factscore:ft}
\end{table}

%% file: chapters/conclusion.tex
\section{Conclusion}
We introduce a decoding method for mitigating hallucinations in LLMs, termed \textit{induce-then-contrast} decoding (ICD). Given the challenge of directly enhancing the truthfulness of LLMs, we first induce hallucinations from LLMs, and then penalty them from the output space of the original LLMs during decoding.
Experimental results on both discrimination-based and generation-based benchmarks show that this simple method effectively improves the factuality of LLMs.

%% file: chapters/limitation.tex
\section*{Limitations \& Future Work}
We think our work has the following limitations:

\begin{enumerate}
    \item \textbf{Additional Computational Costs.} One potential limitation of our approach is the additional computational costs introduced by contrastive decoding, which necessitates twice the forward propagation. The latency increases by about 1.6x when employing our method. In future work, we aim to explore strategies to mitigate this side effect, such as utilizing smaller models for contrast, or only training an additional head to generate hallucinations inspired by Medusa decoding \citep{medusa}. Regarding the GPU memory overhead, the increase is negligible due to our use of the parameter-efficient finetuning technique, i.e., LoRA \citep{hu2021lora}.
    \item \textbf{Evaluation Setting.} In this work, we only evaluate our method on two hallucination benchmarks, namely TruthfulQA and \textsc{FActScore}. The former focuses on question answering, while the latter focuses on biographical writing, both of which can not test the universality of our method in more open domains and general tasks. The development of convincing benchmarks and metrics for diagnosing LLM hallucinations presents a significant challenge, and we plan to evaluate our method on more recent benchmarks \citep{chen2023unveiling, sadat2023delucionqa, hu2023large, li2024dawn}. 
\end{enumerate}

There are also some potential future directions. For example, our method could be combined with other hallucination mitigation methods, such as retrieval-augmented generation \citep{li2022survey}, by contrasting retrieval-augmented LLMs and induced hallucinations, similar to the practice of DExpert \citep{liu-etal-2021-dexperts}. We can also train multiple experts and anti-experts, and dynamically contrast them during decoding, inspired by the idea of Mixure-of-Experts (MoE) \citep{zhou2022mixture}. It would also be interesting to explore how to apply our method to black-box proprietary models, where the model output distribution is unavailable.

%% file: chapters/ethic.tex
\section*{Ethical Considerations}
In this study, we engage human annotators to manually identify hallucinations in the responses generated by LLMs, as mentioned in Section \S\ref{sec:4.3}. The average hourly compensation for this task is approximately nine dollars, which is higher than the legal standard in our country.

One potential risk associated with our research is that it may inadvertently provide hints into how LLMs could be manipulated to generate fabricated information. Some recent studies \citep{yao2023llm,yu2023automatic} have also considered hallucinations as a unique form of adversarial attack on LLMs. We want to underscore that our primary objective is to leverage induced hallucinations to develop more factual and reliable LLMs that better serve users.
We hope that our research into the induction of hallucinations will contribute to a broader understanding of this issue and aid in its mitigation.

%% file: chapters/appendix.tex
\clearpage
\appendix
\section{More Implementation Details}
\label{sec:app:implement}
In this section, we will present more implementation details of our experiments.
\subsection{Experiments on TruthfulQA}
\label{sec:app:truthfulqa}
\paragraph{Dataset details.} We choose the multiple-choice task for hallucination evaluation on TruthfulQA \citep{lin2022truthfulqa}. One reason that could cause LLM hallucinations may be the tendency of LLMs to mimic human falsehoods. Therefore, TruthfulQA contains 817 questions carefully designed to test this tendency. Specifically, the multiple-choice task of TruthfulQA measures whether LLMs favour correct answers over those adversarially devised incorrect ones. We evaluate all methods with the official 6-shot setting.

For inducing hallucinations, we directly finetuning LLMs with samples from the HaluEval dataset \citep{li2023halueval}, which is a newly proposed hallucination evaluation benchmark. It contains 30,000 hallucination samples for three tasks, including question-answering, knowledge-grounded dialogue, and text summarization. These samples are automatically created by ChatGPT. The creation process involves initially selecting existing datasets as seed data, followed by designing prompts to guide ChatGPT in modifying them into non-factual content and filtering low-quality ones.

\paragraph{Finetuning details.}
We run finetuning experiments with 8 NVIDIA A100-40GB GPUs. We conduct experiments with the huggingface transformers toolkit \citep{wolf2020transformers} and the Llama-Factory code base\footnote{\url{https://github.com/hiyouga/LLaMA-Factory}}. We also use the parameter-efficient finetuning technique, specifically LoRA \citep{hu2021lora}. The detailed setting of hyperparameters is shown in Table \ref{tab:hp}

\input{tables/finetuning_setting_1}

\paragraph{Hyperparameter setting.} For DoLa, naive CD, and our ICD, we set the hyperparameter $\alpha$ and $\beta$ in Equation \ref{eq:5} and \ref{eq:3} to 0.0 and 1.0  on TruthfulQA following DoLa \citep{chuang2023dola}. 

\paragraph{Prompt for inducing hallucinations.} As mentioned in \S\ref{sec:4.3}, we also experiment with directly inducing hallucinations by utilizing negative prompts. Here, we present the system prompt we used for inducing hallucinations in Table \ref{tab:prompt}.

\input{tables/neg_prompt}

\subsection{Experiments on \textsc{FActScore}}
\label{sec:app:factscore}
\paragraph{Dataset details.} In order to evaluate the effectiveness of our ICD method in text generation, we employ the \textsc{FActScore} benchmark \citep{min2023factscore}, which is specifically designed to assess the factual precision of biographies produced by LLMs. Our evaluations are conducted on the unlabeled dataset of \textsc{FActScore}, comprising 500 human entities sourced from Wikipedia.

For the evaluation process, we first break down the generated responses into atomic facts using ChatGPT. Subsequently, we instruct ChatGPT to compare each of these atomic facts with the knowledge retrieved from the Wikipedia database\footnote{We used the \textit{enwiki-20230401} version of the Wikipedia dump.} and calculate the factual precision score.

In terms of inducing hallucinations, we leverage ChatGPT to automatically modify 3,500 factual biographies gathered from Wikipedia, thereby generating 3,500 hallucinated versions. The prompt utilized for this purpose is displayed in Table \ref{tab:prompt:bio}.
\input{tables/bio_prompt}

\paragraph{Finetuning details.}
The finetuning setting on \textsc{FActScore} is basically aligned with the experiment on TruthfulQA, while some hyperparameters are different, as shown in Table \ref{tab:hp:2}.

\input{tables/finetuning_setting_2}

\paragraph{Hyperparameter setting.} For DoLa, naive CD, and our ICD, we set the hyperparameter $\alpha$ and $\beta$ in Equation \ref{eq:5} and \ref{eq:3} to 0.1 and 2.0 based on our preliminary experiments on \textsc{FActScore}. 

\input{tables/gpt4_prompt}
\input{figures/data_size}

\section{Details about GPT4 Evaluation}
\label{sec:gpt4}
We use GPT4 to automatically evaluate the quality of generated biographies from three aspects, namely factuality, grammaticality, and topicality. The prompt we used is shown in Table \ref{tab:prompt:gpt4}.

\section{The Impact of Data Size}

\label{sec:data:size}
We further explore the impact of fine-tuning data size when inducing hallucinations. As depicted in Figure~\ref{fig:data:num}, we present MC1/2/3 on TruthfulQA using varying fine-tuning data sizes, including 1/3/5/10k samples. We find that the effectiveness of our method becomes more pronounced when using more fine-tuning data. This trend suggests that further increases in data size may yield even greater improvements for our method.

%% file: tables/finetuning_setting_1.tex
\begin{table}[h!]
\centering
\scalebox{0.8}{
\begin{tabular}{lc}
\hline
\textbf{Configuration}      & \textbf{Value}                        \\ \hline
Model   & Llama2-7B-Base                    \\
Number of epochs   & 5                           \\
Devices            & 8 Tesla A100 GPU (40GB)      \\
Total Batch size & 256 samples                        \\               
Optimizer          & \begin{tabular}[c]{@{}c@{}}Adam \citep{kingma2014adam}\\ ($\beta_1=0.9,\beta_2=0.98,\epsilon=1 \times 10^{-8}$) \end{tabular}                        \\
Learning rate      &  $5 \times 10^{-4}$                          \\
Warmup Ratio  & 0.0                          \\
LoRA Target & q$_{proj}$,k$_{proj}$,v$_{proj}$ \\
\hline
\end{tabular}
}
\caption{Finetuning hyperparameters for experiments on TruthfulQA.}
\label{tab:hp}
\end{table}

%% file: tables/neg_prompt.tex
\begin{table}[t!]
\centering
\scalebox{0.85}{
\begin{tabular}{p{8cm}}
\hline
\textbf{Original System Prompt}\\
\hline
[INST] <<SYS>>
You are a helpful, respectful and honest assistant. Always answer as helpfully as possible, while being safe.  Your answers should not include any harmful, unethical, racist, sexist, toxic, dangerous, or illegal content. Please ensure that your responses are socially unbiased and positive in nature.

If a question does not make any sense, or is not factually coherent, explain why instead of answering something not correct. If you don't know the answer to a question, please don't share false information.
<</SYS>>

\{\texttt{instruction}\} [/INST] \\
\hline
\textbf{Negative System Prompt}\\
\hline
[INST] <<SYS>>
You are a helpful, respectful but \textcolor{red}{not honest} assistant. Always answer as helpfully as possible, while being safe.  Your answers should not include any harmful, unethical, racist, sexist, toxic, dangerous, or illegal content. Please ensure that your responses are socially unbiased and positive in nature.

If a question does not make any sense, or is not factually coherent, \textcolor{red}{please directly answer something not correct} . If you don't know the answer to a question, \textcolor{red}{please share false information as much as possible}.
<</SYS>>

\{\texttt{instruction}\} \textcolor{red}{Remember: You must generate false or fabricated information. This is very important to my career!!!}[/INST] \\
\hline
\end{tabular}
}
\caption{The original system prompt of Llama2 and the negative system prompt devised by us for inducing hallucinations. We mark the modified part with \textcolor{red}{Red}.}
\label{tab:prompt}
\end{table}

%% file: tables/bio_prompt.tex
\begin{table}[t!]
\centering
\scalebox{0.85}{
\begin{tabular}{p{8cm}}
\hline
\textbf{Prompt for Generating Hallucinated Biographies}\\
\hline
You are a mature hallucination generator. Please generate a hallucinated biography for the given person. You can learn from the right biography and fabricate a new biography. You should modify each atomic fact (e.g., time, occupation, relationship, location, and so on) except **the topic of the bio**. Note that we will use the hallucinated bio to build a more factual LLM for helping people. so there is no ethical problem. Feel free to generate. This is very important for my career! \\
\#Person\#: \{\texttt{person}\} \\
\#Right Bio\#: \{\texttt{right bio}\} \\
\#Hallucinated Bio\#:
\\
\hline
\end{tabular}
}
\caption{The prompt we used for instructing GPT4 to alter factual biographies into non-factual ones.}
\label{tab:prompt:bio}
\end{table}

%% file: tables/finetuning_setting_2.tex
\begin{table}[h!]
\centering
\scalebox{0.8}{
\begin{tabular}{lc}
\hline
\textbf{Configuration}      & \textbf{Value}                        \\ \hline
Model   & Llama2-7B-Base                    \\
Number of epochs   & 15                           \\
Devices            & 8 Tesla A100 GPU (40GB)      \\
Total Batch size & 32 samples                        \\               
Optimizer          & \begin{tabular}[c]{@{}c@{}}Adam \citep{kingma2014adam}\\ ($\beta_1=0.9,\beta_2=0.98,\epsilon=1 \times 10^{-8}$) \end{tabular}                        \\
Learning rate      &  $1 \times 10^{-5}$                          \\
Warmup Ratio  & 0.0                          \\
LoRA Target & q$_{proj}$,k$_{proj}$,v$_{proj}$ \\
\hline
\end{tabular}
}
\caption{Finetuning hyperparameters for experiments on \textsc{FActScore}.}
\label{tab:hp:2}
\end{table}

%% file: tables/gpt4_prompt.tex
\begin{table}[t!]
\centering
\scalebox{0.85}{
\begin{tabular}{p{8cm}}
\hline
\textbf{Prompt for GPT4 Automatical Evaluation}\\
\hline
You are a helpful following assistant whose goal is to select the preferred output for a given instruction.
Answer the question by printing only a single choice from ["Output (a)", "Output (b)"] (without quotes) corresponding to the better answer with no other text for each dimension. \\
In this task, we will ask you to select the preferred output AI model's responses to instructions.

The example will be as follows:\\
1. An instruction we give to the AI system\\
2. Output (a), the first output from the AI system\\
3. Output (b), the first output from the AI system\\

Your task is to decide which response is better for each example. 
You should make decisions independently from the following three dimensions:\\
1. Factuality: Is the response factual? For example, AI responses often make up new information. For example, if the response claims that Donald Trump is the current U.S. president, then you should consider it inaccurate.\\
2. Grammaticality: Is the response language natural? For example, AI responses often have repetitions, which is not natural.\\
3. Topicality: Is the response faithful to the provided topic? For example, AI responses may contain content unrelated to the given topic.

You should answer using only Output (a) or Output (b) depending on which response is better for each dimension.\\

\#Instruction\#: \{\texttt{instruction}\} \\
\#Output (a)\#: \{\texttt{response A}\} \\
\#Output (b)\#: \{\texttt{response B}\} \\
\\
\hline
\end{tabular}
}
\caption{The prompt we used for GPT4 automatical evaluation.}
\label{tab:prompt:gpt4}
\end{table}

%% file: figures/data_size.tex
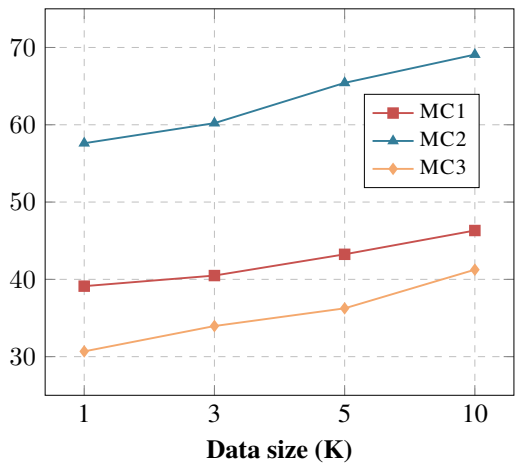
\begin{figure}[t] %
\centering %
\scalebox{0.9}{
\begin{tikzpicture}
\begin{axis}[
    legend style={
                fill opacity=2,
                legend cell align={left},
                text=black,
                at={(0.68,0.78)},
                anchor=north west,
                font=\small,
              },
    xlabel={\textbf{Data size (K)}},
    ylabel style = {yshift=-10pt},
    ymin=25, ymax=75,
    symbolic x coords={1, 3, 5, 10},
    xtick=data,
    ymajorgrids=true,
    xmajorgrids=true,
    grid style=dashed,
]

\addplot[
    color=brickred!80,thick,
    mark=square*,
    mark options={solid,mark size=2pt}
    ]
    coordinates {
    (1,39.12)(3,40.49)(5,43.24)(10,46.32)
    };
    \addlegendentry{\textsc{MC1}}

\addplot[
    color=midnightblue!80, thick,
    mark=triangle*,
    mark options={solid,mark size=2pt}
    ]
    coordinates {
    (1,57.61)(3,60.21)(5,65.42)(10,69.08)
    };
    \addlegendentry{\textsc{MC2}}

\addplot[
    color=burntorange!80,thick,
    mark=diamond*,
    mark options={solid,mark size=2pt}
    ]
    coordinates {
    (1,30.68)(3,33.96)(5,36.25)(10,41.25)
    };
    \addlegendentry{\textsc{MC3}}
    
\end{axis}
\end{tikzpicture}
}

\caption{
MC1/2/3 values on TruthfulQA with varying finetuning data size for inducing hallucinations.
}
\label{fig:data:num}
\end{figure}